\pdfoutput=1
\documentclass[12pt]{article}
\usepackage{do-calc}
\begin{document}
\title{Introduction to\\
Judea Pearl's Do-Calculus}

\author{Robert R. Tucci\\
        P.O. Box 226\\
        Bedford,  MA   01730\\
        tucci@ar-tiste.com}

\date{\today}
\maketitle
\vskip2cm
\section*{Abstract}
This is a purely
pedagogical paper
with no new results. The goal
of the paper is to
give a fairly self-contained
 introduction to Judea Pearl's
do-calculus, including proofs of his
3 rules.
\newpage
\section{Introduction}
Judea Pearl's do-calculus is
a part of his theory of probabilistic
causality, which itself is a part
of the study of Bayesian networks
(for which he is largely responsible too).
For a
good textbook on Bayesian Networks,
see, for example, Ref.\cite{KF}
by Koller and Friedman.

The goal of this paper
is to give a fairly self-contained
introduction to Judea Pearl's
do-calculus.
Pearl first enunciated
his calculus in
the 1995 paper Ref.\cite{P95}.
Our paper is mostly based on
Ref.\cite{P95}.
Compared with
Ref.\cite{P95},
the scope of our paper is
smaller (for example,
we don't discuss
 ``identifiability" at all).
However,
 for the material we do cover,
 we present some extra
 details which are not found
 in Ref.\cite{P95} and
which might be helpful to beginners.
Ref.\cite{P95} is a wonderful paper
and we fully expect our readers
to read it at the same time that they read
this one. We just
think that it might help
the readers of
Ref.\cite{P95} to hear the same thing
explained by someone else,
in slightly different words, and
from a slightly different perspective.

In this paper,
we give full proofs of the
3 rules of do-calculus.
Our proofs are
almost the same but
slightly different from those
found
in the Appendix of
Ref.\cite{P95}.

Since 1995,
some interesting new
consequences,
ramifications and
applications of
Pearl's do-calculus
have been found. These
were reviewed recently (2012)
by
Pearl in Ref.\cite{P12}.

\section{Basic Notation}

In this section,
we will define some basic notation
that will be used
later in the paper.

We will use $\delta_a^b$ or
$\delta(a,b)$ to
denote the Kronecker delta function (equals
1 if $a=b$ and 0 otherwise).

We will indicate random variables
by underlined symbols
and indicate their possible values (a.k.a. states,
instances)
by the same letter, but not underlined.
For example, $\rva$ takes on values $a$.
Many people, Pearl
and coworkers included,
indicate random
variables by capital letters
and their possible values by
lower case letters.
For example, $A$ takes on values $a$.

Given a probability distribution
$P_{\rva,\rvb}(a,b)$,
let

\beq
P(a:b)=\frac{P(a,b)}{P(a)P(b)}=
\frac{P(a|b)}{P(a)}
\;,
\eeq
and

\beq
P(a:b|c)=\frac{P(a,b|c)}{P(a|c)P(b|c)}=
\frac{P(a|b,c)}{P(a|c)}
\;.
\eeq

We will indicate
n-tuples (vectors, ordered sets) by a
 letter followed by a
dot, as in $x.=(x_1,x_2, \ldots,x_n)$.
 The dot is intended
to suggest
that the subscript is free.
Many people denote n-tuples by putting an
arrow over the letter (as in $\vec{x}$)
or by using a boldface letter
(an in $\bf{x}$).

Often, we will treat two n-tuples of
random variables as
if they were plain sets and
use them in conjunction with standard
set symbols such as those for
subset, union, intersection
and subtraction. For example,
if $\rvx.$ and $\rvy.$
are an n-tuple and an m-tuple, respectively,
where $m$ and $n$ are not
necessarily the same,
then we might write
$\rvx.\subset \rvy.$, $\rvx.\cup\rvy.$,
$\rvx.\cap\rvy.$ and $\rvx.-\rvy.$.
In such contexts, we will sometimes
not distinguish between $\rvx_j$ and
the singleton set $\{\rvx_j\}$. For example
we might write $\rvx.-\rvx_j$ instead of
$\rvx.-\{\rvx_j\}$.

A classical Bayesian network is
a
DAG (directed acyclic graph)
where each vertex (a.k.a. node) is labeled
by a random variable $\rvx_j$ and
is assigned a
transition  probability matrix
about which we will say more below.
Let $\rvx. =
(\rvx_1,\rvx_2,\ldots,\rvx_N)$.
Arrows are also called directed edges.
Each node $\rvv$ with an arrow
going from $\rvv$ to $\rvx_j$
is called a
parent of $\rvx_j$ and
the set of such
parent nodes is denoted by $\rv{pa}(\rvx_j)$.
Each node $\rvv$ with an arrow
going from $\rvx_j$
to $\rvv$
is called a
child of $\rvx_j$ and
the set of such
children nodes is denoted
by $\rv{ch}(\rvx_j)$.
Each node $\rvx_j$ is assigned
a transition  probability matrix
$P(x_j|pa(\rvx_j))$
that depends on the value
$x_j$ of node $\rvx_j$
and the values $pa(\rvx_j)$
of nodes $\rv{pa}(\rvx_j)$.
The entire Bayesian network is assigned a total
probability

\beq
P(x.)= \prod_{j=1}^{N}
P(x_j|pa(\rvx_j))
\;.
\eeq

\section{Subgraphs and Augmented Graphs}

In this section, we will
define certain subgraphs
and augmented graphs,
derived from a graph $G$, that
will be especially useful
to us later on.

Suppose that graph $G$
has nodes $\rvx.$ and $\rva.\subset \rvx.$.

\begin{figure}[h]
    \begin{center}
    \includegraphics[height=2.0in]{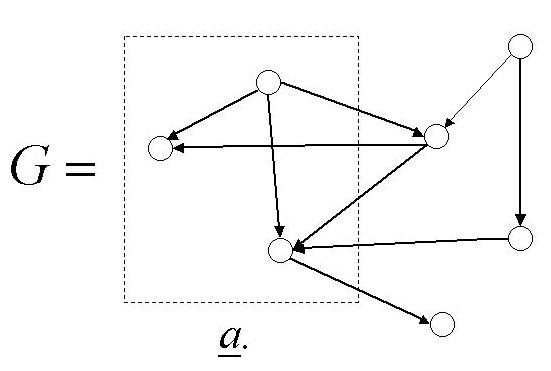}
    \caption{A corral for subset $\rva.$
    of the nodes $\rvx.$ of
    graph $G$.
    }
    \label{fig-corral}
    \end{center}
\end{figure}
We can draw a frame
that encloses
the nodes $\rva.$ and leaves
the nodes $\rvx.-\rva.$ outside.
We will
refer to
such an enclosure as the
$\rva.$ ``corral" (See Fig.\ref{fig-corral}
for an example).

\begin{figure}[h]
    \begin{center}
    \includegraphics[height=2.0in]{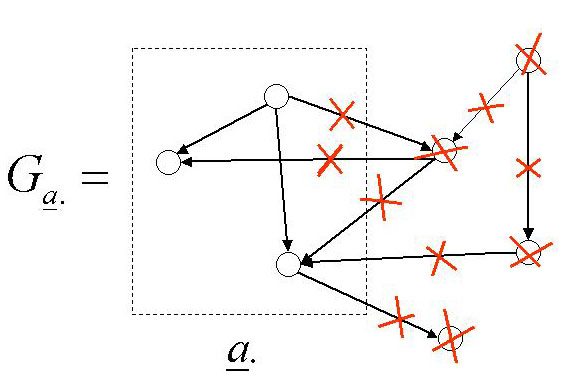}
    \caption{Graph $G_{\rva.}$
    arises from
    graph $G$ of Fig.\ref{fig-corral}
    if we restrict $G$
    to node set $\rva.$.
    Arrows or nodes with a red cross
    through them should be erased.
    }
    \label{fig-restriction}
    \end{center}
\end{figure}
We will use $G_{\rva.}$
to denote the ``restriction" of
graph $G$ wherein
nodes $\rvx.-\rva.$
and any arrows connected to
$\rvx.-\rva.$ have
been erased.
(See Fig.\ref{fig-restriction}
for an example).

\begin{figure}[h]
    \begin{center}
    \includegraphics[height=2.0in]{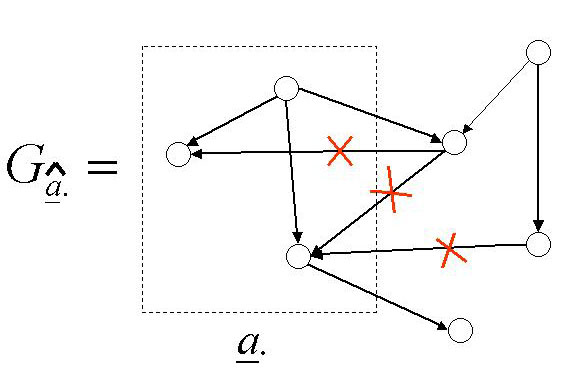}
    \caption{Graph $G_{\myhat{\rva}.}$
    arises from graph $G$ of Fig.\ref{fig-corral}
    if we erase
    from $G$ all arrows entering node set $\rva.$.
    Arrows or nodes with a red cross
    through them should be erased.
    }
    \label{fig-in-ban}
    \end{center}
\end{figure}
We will use $G_{\myhat{\rva}.}$
to denote the graph $G$ with
arrows
entering $\rva.$ erased.
Mnemonic: Think of the ``hat" on top of $\rva.$
as being the arrow-head of
an arrow exiting $\rva.$.
Only arrows of this type (that is, those
that
are exiting $\rva.$)
are allowed.(See Fig.\ref{fig-in-ban}
for an example).

\begin{figure}[h]
    \begin{center}
    \includegraphics[height=2.0in]{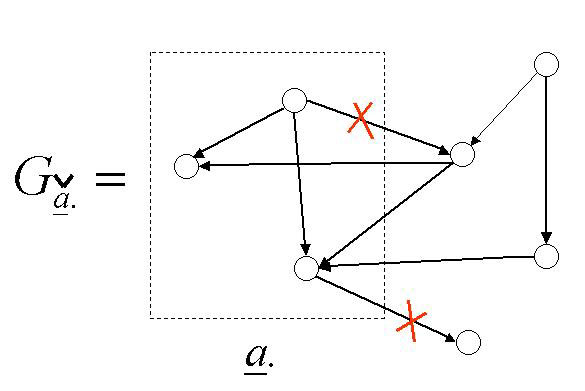}
    \caption{Graph $G_{\myvee{\rva}.}$
    arises from graph $G$ of Fig.\ref{fig-corral}
    if we erase from $G$
     all arrows exiting node set $\rva.$.
    Arrows or nodes with a red cross
    through them should be erased.
    }
    \label{fig-out-ban}
    \end{center}
\end{figure}
We will use $G_{\myvee{\rva}.}$
to denote the graph $G$ with
arrows
exiting $\rva.$ erased.
Mnemonic: Think of the ``vee" on top of $\rva.$
as being the arrow-head of
an arrow entering $\rva.$.
Only arrows of this type (that is, those
that
are entering $\rva.$)
are allowed.(See Fig.\ref{fig-out-ban}
for an example).

\begin{figure}[h]
    \begin{center}
    \includegraphics[height=2.0in]{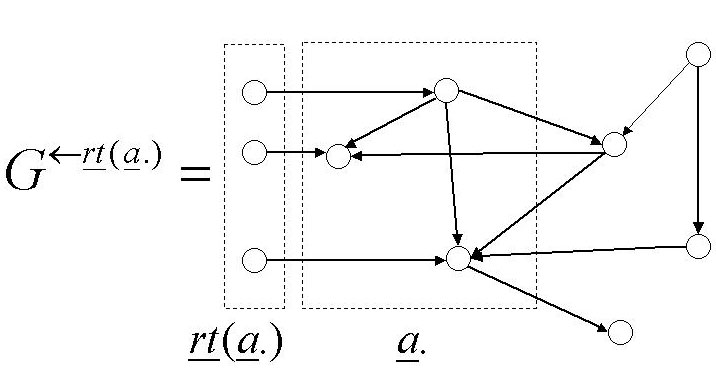}
    \caption{Graph $G\larrow\rv{rt}(\rva.)$
    arises from graph $G$ of Fig.\ref{fig-corral}
    if add to $G$
    a node set $\rv{rt}(\rva.)$
    of root nodes for $\rva.$.
    }
    \label{fig-add-rt}
    \end{center}
\end{figure}

We will use $G\larrow \rv{rt}(\rva.)$
to denote
the augmented graph
obtained by adding to graph $G$
 a node set
$\rv{rt}(\rva.)$
and arrows
from $\rv{rt}(\rva.)$
to node set $\rva.$
 ($\rva.$ is contained in $G$).
For each $\rva_j\in \rva.$,
one adds exactly one
twin node
$\rv{rt}(\rva_j)\in
\rv{rt}(\rva.)$,
and one arrow from
the ``root" node
$\rv{rt}(\rva_j)$ to
$\rva_j$.(See Fig.\ref{fig-add-rt}
for an example).

\begin{figure}[h]
    \begin{center}
    \includegraphics[height=2.0in]{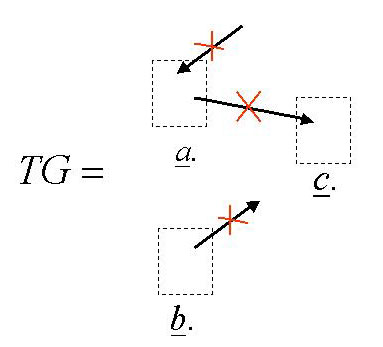}
    \caption{Example
    of a graph template $TG$.
    This one has a
     ban on arrows entering $\rva.$,
    a ban on arrows exiting $\rvb.$,
    and a ban on arrows going
    from $\rva.$ to $\rvc.$.
    }
    \label{fig-g-template}
    \end{center}
\end{figure}

One can
describe a set or
family of graphs by
 using what I call
 a graph template.
In a graph template,
some corrals
have bans or restrictions on
the types of arrows
that are allowed to
cross the fence.
A ban is represented
by an arrow with a red cross
on it to indicate the
type of arrow that is forbidden.
One can ban this way either
all arrows entering the corral
or all arrows exiting the corral
or all arrows going from one
corral to another.
In other words,
if nodes are like cows, some corrals
have one way gates that ban
certain types of bovine movement.
See Fig.\ref{fig-g-template}
for an example of a graph template $TG$.

For $f\in\{ch,de,pa,an\}$ and $\rvv.\subset \rvx$, let
$f(\rva., G_{\rvv.})$ be the set of
nodes which are the
children, descendants, parents
and ancestors, respectively, of $\rva.$ in
the graph $G_{\rvv.}$.
We will write $f(\rva.)$ instead
of $f(\rva., G_{\rvv.})$
when $G_{\rvv.}=G_{\rvx.}=G$.
If $f^{(n)}(\cdot)$ indicates
application $n$ times of
the function $f(\cdot)$,
then
$de(\rva.,G_{\rvv.})= \cup_{n=1}^{\infty}
ch^{(n)}(\rva.,G_{\rvv.})$
and
$an(\rva.,G_{\rvv.})= \cup_{n=1}^{\infty}
pa^{(n)}(\rva.,G_{\rvv.})$.
For $f\in\{ch,de,pa,an\}$,
let
$\overline{f}(\rva.,G_{\rvv.})=f(\rva.
,G_{\rvv.})\cup\rva.$
and call $\overline{f}(\cdot)$
the closure of $f(\cdot)$.

\section{D-Separation}

In this section,
we will explain
the d-sep (dependence separation)
theorem, which
tells us how to
diagnose from a graph $G$
whether $\rva.$
and $\rvb.$ are
probabilistically conditionally
independent at
fixed $\rve.$,
where $\rva.,\rvb.,\rve.$
are disjoint subsets of
nodes of $G$.

\begin{figure}[h]
    \begin{center}
    \includegraphics[height=1.5in]{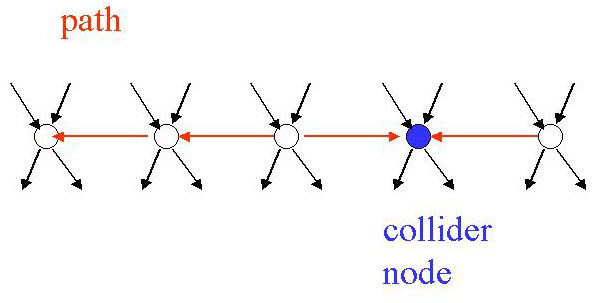}
    \caption{A typical path of a graph and
    a typical collider node in that path.
    }
    \label{fig-path}
    \end{center}
\end{figure}

\begin{figure}[h]
    \begin{center}
    \includegraphics[height=3.0in]{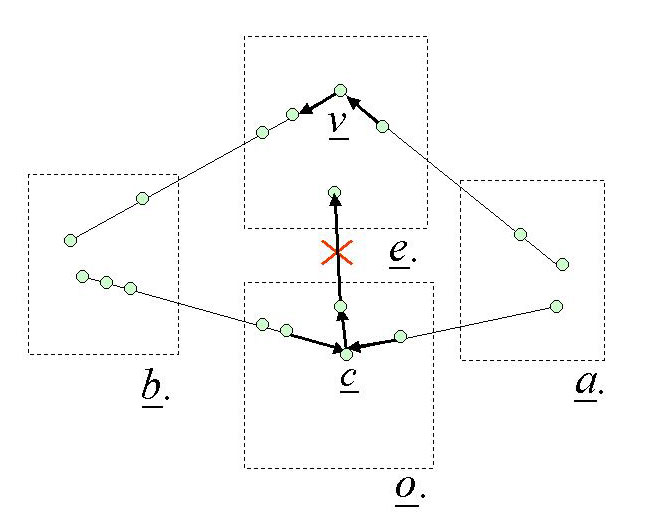}
    \caption{Two types of paths
    from $\rva.$ to $\rvb.$
    that
    are blocked
    at fixed $\rve.$. The set
    $\rvo.$ is defined to contain
    all ``other" nodes; i.e., all nodes
    not in $\rva.\cup\rvb.\cup\rve.$.
    Note that the collider node $\rvc$
    can have descendants
    in either $\rvb.,\rvo.$ or $\rva.$.
    Note also that even though in this figure
    we put $\rvc$ in the $\rvo.$ corral,
    it could also be in $\rvb.$ or $\rva.$.
    }
    \label{fig-d-sep}
    \end{center}
\end{figure}

Suppose that graph $G$
has nodes $\rvx.$ and $\rva.\subset \rvx.$.
If all the nodes in $\rva.$
are like the beads in
a beaded string
with one arrow
between adjacent beads,
where the direction of the
arrows may change inside the string,
then we will call $\rva.$
an undirected path of $G$.

We will use $Path_G(\rv{A}<\rv{B})$
to denote the set of all
undirected paths
in graph $G$ that
start at node $\rv{A}$ and end at
node $\rv{B}$. Here $<$ means
that there are $\geq 1$ arrows
(in whatever direction)
and $\geq 0$  nodes between
$\rv{A}$ and $\rv{B}$.
We will also use
$\rv{A}\leq \rv{B}$
if $\rv{A}$ and $\rv{B}$
could be the same node.
We will also write
a comma instead of a $<$
between the $\rv{A}$ and $\rv{B}$
if there is only one arrow
between $\rv{A}$ and $\rv{B}$.
If $\rva.$ and $\rvb.$
are disjoint subsets of $\rvx.$,
let $Path_G(\rva.<\rvb.)=\cup_{\rv{A}\in\rva.,
\rv{B}\in \rvb.}Path(\rv{A}<\rv{B})$.
If $\rva.^{(1)}, \rva.^{(2)},\ldots,\rva.^{(n)}$
are disjoint subsets of $\rvx.$,
define
$Path_G(\rva.^{(1)}< \rva.^{(2)}<\ldots<\rva.^{(n)})$
as the obvious generalization of
this notation.

Given an undirected path $\gamma$ of $G$,
any node which has arrows impinging upon it
from both sides of the string will be called
a collider node of $\gamma$.
(See Fig.\ref{fig-path} for an example).
We will denote the set of
all collider nodes of path $\gamma$
by $col(\gamma)$.

Suppose a graph $G$
has nodes $\rvx.$ and that
$\rvx.$ equals the union of
the disjoint sets
$\rva,\rvb.,\rve.,$ and $\rvo.$.
$\gamma \in Path(\rva.<\rvb.)$
is said to be
\textbf{blocked }
at fixed $\rve.$ if either
\begin{itemize}
\item
$(\exists \rvv\in \gamma)
[\rvv\notin col(\gamma)
\mbox{ and }
\rvv\in \rve.$], or
\item
$(\exists \rvc\in \gamma)
[\rvc\in col(\gamma)
\mbox{ and }
\overline{de}(\rvc)
\cap\rve.=\emptyset]$.

See Fig.\ref{fig-out-ban}
for a picture of
these two types of blocked paths.
I like to think of the non-collider
 node $\rvv$ as a canyon pass which is
blocked by an obstacle (like a boulder)
in $\rve.$
thus impeding information
and cattle from
flowing through the path.
As for the collider node $\rvc$,
I like to think of it as a
deep sink-hole that
is an obstacle to cattle.
However, if sink-hole $\rvc$
is in $\rve.$ or
even if merely one of
its descendants
is in $\rve.$,
then this has the effect of
filling that sink-hole
so that information
and cattle can once
again flow through the path.
\end{itemize}

By negating the previous definition,
we immediately get that
$\gamma \in Path(\rva.<\rvb.)$
is
\textbf{unblocked }
at fixed $\rve.$ if
\begin{itemize}
\item
$(\forall \rvv\in \gamma)[
\rvv\notin col(\gamma)
\implies
\rvv\notin \rve.]$, and
\item
$(\forall \rvc\in \gamma)[
\rvc\in col(\gamma)
\implies
\overline{de}(\rvc)\cap\rve.
 \neq \emptyset]$.
\end{itemize}

 We write
 $(\rva.\bot\rvb.|\rve.)_G$
and read this as $\rva.$
and $\rvb.$ are \textbf{d-sep}
 (dependance-separated)
 at fixed $\rve.$
 in graph $G$ if
all $\gamma\in Path_G(\rva.<\rvb.)$
are blocked at fixed $\rve.$.

\begin{claim}(D-Sep Theorem):
\newline
$(\rva.\bot\rvb.|\rve.)_G$
if and only if, for all possible values of
$a.,b.,e.$,
$P_G(a.:b.|e.)=1$,
or, equivalently,
$P_G(a.|b.,e.)=P_G(a.|e.)$.
\end{claim}
\proof
See Ref.\cite{P95}
for a history of this theorem,
including pertinent references.
Ref.\cite{P95}
also describes and gives references for an
alternative graphical
method, invented
by Lauritzen, of diagnosing
d-sep.
\qed

\section{Uprooting And Mowing a Node}

In this section,
we will define
two operations for pruning the arrows
connected to a node.
One operation ``uproots the node",
meaning that it erases all the roots
(i.e., incoming
arrows) of the node.
The other
``mows the node",
meaning that it erases all the
stems (i.e., outgoing
arrows) of the node.
Uprooting a node is called
an ``intervention"
by Pearl and co-workers.

Through out this section, let
$G$
be a graph with nodes $\rvx.$
and let $\rva.,\rvb.,\rve.$ be
3 disjoints
subsets of $\rvx.$.
\subsection{Definitions}

\begin{itemize}
\item {\bf Uprooting}

We define
as follows
the probability
that $\rvb.=b.$ when
$\rva.=a.$ is uprooted:

\beq
P(b.|\myhat{a}.) =
\frac{P_{G_{\myhat{\rva}.}}(a.,b.)}
{P_{G_{\myhat{\rva}.}}(a.)}
\neq P_G(b.|a.)
\;.
\eeq
Here $P_{G_{\myhat{\rva}.}}(x.)$
is the probability distribution
for the subgraph $G_{\myhat{\rva}.}$
of $G$.
$P_{G_{\myhat{\rva}.}}(x.)$
 is defined from $P(x.)$
 by replacing $P(a_j|pa(\rva_j))$
 by $P(a_j)$ for all $j$.
 Note that
 when we do this replacement,
 all the
arrows entering $\rva.$ (the
``roots" of $\rva.$) are
being erased or ``severed".

Other notations
used in the literature for
$P(b.|\myhat{a}.)$
are $P(b.|do(\rva.)=a.)$
(where $do(\cdot)$ is called
the do operator), and
$P_{a.}(b.)$.
We will sometimes write
$[{\cal S}]^\wedge$ instead
of $\myhat{{\cal S}}$,
especially when ${\cal S}$
is a long expression.

An equivalent
definition of $P(b.|\myhat{a}.)$
is as follows. We define

\beqa
P(x.-a.|\myhat{a}.)
&=&
\frac{P(x.)}
{\prod_{j:\rvx_j\in \rva.}
P(x_j|pa(\rvx_j))}
\label{eq-uprooting}
\\
&=&
\prod_{j:\rvx_j\in (\rvx.-\rva.)}
P(x_j|pa(\rvx_j))
\;.
\eeqa
Note that
$\sum_{x.-a.}
P(x.-a.|\myhat{a}.)=1$.
Note also that
if $\rva.=(\rva_1,\rva_2,\ldots,\rva_n)$,
then $P(x.-a.|\myhat{a}.)=
P(x.-a.|\myhat{a}_1,
\myhat{a}_2,\ldots,\myhat{a}_n)$.

Next we define

\beq
P(b.|\myhat{a}.)=
\sum_{x.-(b.\cup a.)}
P(x.-a.|\myhat{a}.)
\;.
\eeq
We also define

\beq
P(b.|\myhat{a}.,e.)=
\frac{
P(b.,e.|\myhat{a}.)
}{
P(e.|\myhat{a}.)
}
\;.
\eeq

I like to call
$P(b.|\myhat{a}.,e.)$
the probability of $\rvb.$ conditioned
on $\rve.$, and with $\rva.$ uprooted.

Yet another equivalent
definition of $P(b.|\myhat{a}.)$
is as follows. For this definition,
 we begin by
augmenting the graph $G$ to
$G\larrow \rv{rt}(\rva.)$.
In the new graph,
each node $\rva_j$ has
a new incoming arrow.
We define the transition
matrices for the nodes $\rva_j$
in the new graph from the
transition matrices
of the old graph as follows.
For all $j$
and for all
values $a_j$ of
 $\rva_j$, let

\beq
P(a_j | pa(\rva_j, G), rt(\rva_j))=
P(a_j)\delta_{rt(\rva_j)}^{1}+
P(a_j | pa(\rva_j, G))\delta_{rt(\rva_j)}^{0}
\;.
\eeq
Note that

\beq
\left\{
\begin{array}{l}
P(b.|a.,rt(\rva.)=0)=P(b.|a.)\\
P(b.|a.,rt(\rva.)=1)=P(b.|\myhat{a}.)
\end{array}
\right.
\;,
\eeq
where $rt(\rva.)=n$ for $n\in\{0,1\}$
means
$rt(\rva_j)=n$ for all $j$.
Thus, node set $\rv{rt}(\rva.)$
acts like a switch. When it is on
(i.e., when it equals 1),
all the roots of node set $\rva.$
are severed.

\item {\bf Mowing}

Note that

\beqa
P(x.)&=&\prod_j P(x_j|pa(\rvx_j))\\
&=&
\prod_{j:\rvx_j\in(\rvx.-\rva.)}
\left\{P(x_j|pa(\rvx_j))\right\}
\prod_{j:\rvx_j\in\rva.}
\left\{P(x_j|pa(\rvx_j))\right\}\\
&=&
P(x.-a.|[a.]^\wedge)
P(a.|[x.-a.]^\wedge)
\;.
\eeqa
We define as follows
the probability
that $\rvx.=x.$ when
$\rva.=\ap$ is mowed:

\beqa
P_{\myvee{\rva}.(\ap)}(x.)&=&
\left[P(x.-a.|[a.]^\wedge)
\right]_{a.\rarrow \ap}
P(a.|[x.-a.]^\wedge)\\
&=&
P(x.-a.|[\ap]^\wedge)
P(a.|[x.-a.]^\wedge)
\;.
\label{eq-mow-x}
\eeqa
Note that
we set to $\ap$
the value of $\rva.$
at the destinations of the
arrows exiting $\rva.$.
By doing this, we are severing
the outgoing arrows of $\rva.$.
Note that
$\sum_{x.}
P_{\myvee{\rva}.(\ap)}(x.)=1$
and
$P_{\myvee{\rva}.(\ap)}(x.)=
P_{\prod_j\myvee{\rva}_j(a_j')}(x.)$.

Next we define

\beq
P_{\myvee{\rva}.(\ap)}(a.,b.)
\sum_{x.-(a.\cup b.)}
P_{\myvee{\rva}.(\ap)}(x.)
\;.
\eeq

We also define

\beq
P_{\myvee{\rva}.(\ap)}(b.|a.,e.)=
\frac{
P_{\myvee{\rva}.(\ap)}(a.,b.,e.)
}{
P_{\myvee{\rva}.(\ap)}(a.,e.)
}
\;.
\eeq

Note that summing both sides of
Eq.(\ref{eq-mow-x}) over $a.$
 yields

\beq
P_{\myvee{\rva}.(\ap)}(x.-a.)
=
P(x.-a.|[\ap]^\wedge)
\;,
\label{eq-mow-x-a}
\eeq
and summing both sides of
Eq.(\ref{eq-mow-x-a})
over $x.-(a.\cup b.)$ yields

\beq
P_{\myvee{\rva}.(\ap)}(b.)
=
P(b.|[\ap]^\wedge)
\;.
\eeq

\end{itemize}
\subsection{Operators}

Let $pd(\rvx.)$
be the set of all
possible probability distributions
for $\rvx.$, where $\rvx.$
labels the nodes of the graph $G$.
Let ${\cal L}(pd(\rvx.))$
denote the set of all linear
combinations over the reals of the
elements of  $pd(\rvx.)$.
It is convenient to define
linear operators acting on
${\cal L}(pd(\rvx.))$
whose effect is to mow and
uproot a node.

Let

\beq
Cond_{\rva.} P(a.,b.) =
P(b.|a.)
\;.
\eeq

\begin{itemize}

\item {\bf Uprooting}

We define
as follows a linear
operator $\delta_{\myhat{\rva}.}$
that does uprooting of $\rva.$

\beq
\delta_{\myhat{\rva}.}P(a.,b.)=
P(b.|\myhat{a}.)
\;.
\label{eq-uproot-prop1}
\eeq
Note that
$\delta_{\myhat{\rva}.}=
\prod_j \delta_{\myhat{\rva}_j}$
and $\delta_{\myhat{\rva}.}P(a.)=1$.
Next we extend
the domain of
$\delta_{\myhat{\rva}.}$
as follows so that, besides
acting on ${\cal L}(pd(\rvx.))$,
it can also act
on a ratio of two elements of
${\cal L}(pd(\rvx.))$.

\beq
\delta_{\myhat{\rva}.}
P(b.|a.,e.)=
\frac{
\delta_{\myhat{\rva}.}
P(a.,b.,e.)
}{
\delta_{\myhat{\rva}.}
P(a.,e.)
}
\;.
\label{eq-uproot-prop2}
\eeq

\begin{claim}
\beq
\delta_{\myhat{\rva}.}
P(a.,b.)
=P(b.|\myhat{a}.)
\;,
\eeq

\beq
\delta_{\myhat{\rva}.}
P(b.)
=\sum_{a.}P(b.|\myhat{a}.)
\;,\;\;\left[
\delta_{\myhat{\rva}.}\sum_{a.}=
\sum_{a.}\delta_{\myhat{\rva}.}
\right]
\;,
\eeq

\beq
\delta_{\myhat{\rva}.}
P(b.|a.)
=P(b.|\myhat{a}.)
\;,\;\;\left[
\delta_{\myhat{\rva}.}Cond_{\rva.}=
\delta_{\myhat{\rva}.}
\right]
\;.
\eeq
\end{claim}
\proof
This all follows easily from
the linearity of $\delta_{\myhat{\rva}.}$
and Eqs.(\ref{eq-uproot-prop1}) and
 (\ref{eq-uproot-prop2}).
\qed

\item {\bf Mowing}

We define
as follows a linear
operator $\delta_{\myvee{\rva}.(\ap)}$
that does mowing of $\rva.$
to $\ap$

\beq
\delta_{\myvee{\rva}.(\ap)}
P(a.,b.)
=
P_{\myvee{\rva}.(\ap)}(a.,b.)
\;.
\label{eq-mow-prop1}
\eeq
Note that
$\delta_{\myvee{\rva}.(\ap)}=
\prod_j\delta_{\myvee{\rva}_j(a'_j)}$
and
$\delta_{\myvee{\rva}.(\ap)}P(a.)=
P(a.)$.
Next we extend the domain of
$\delta_{\myvee{\rva}.(\ap)}$,
as follows so that
it can also act
on a ratio of two elements of
${\cal L}(pd(\rvx.))$.

\beq
\delta_{\myvee{\rva}.(\ap)}
P(b.|a.,e.)=
\frac{
\delta_{\myvee{\rva}.(\ap)}
P(a.,b.,e.)
}{
\delta_{\myvee{\rva}.(\ap)}
P(a.,e.)
}
\;.
\label{eq-mow-prop2}
\eeq

\begin{claim}
\beq
\lim_{\ap\rarrow a.}
\delta_{\myvee{\rva}.(\ap)}
P(a.,b.)
=P(a.,b.)
\;,\;\;\left[
\lim_{\ap\rarrow a.}
\delta_{\myvee{\rva}.(\ap)}=1
\right]
\;,
\eeq

\beq
\lim_{\ap\rarrow a.}
\delta_{\myvee{\rva}.(\ap)}
P(b.)
=P(b.|\myhat{a}.)
\;,\;\;\left[
\lim_{\ap\rarrow a.}
\delta_{\myvee{\rva}.(\ap)}
\sum_{a.}=
\delta_{\myhat{\rva}.}
\right]
\;,
\eeq

\beq
\lim_{\ap\rarrow a.}
\delta_{\myvee{\rva}.(\ap)}
P(b.|a.)
=P(b.|a.)
\;,\;\;\left[
\lim_{\ap\rarrow a.}
\delta_{\myvee{\rva}.(\ap)}
Cond_{\rva.}=
Cond_{\rva.}
\right]
\;.
\eeq
\end{claim}
\proof
This all follows easily from
the linearity of $\delta_{\myvee{\rva}.(\ap)}$
and Eqs.(\ref{eq-mow-prop1}) and
 (\ref{eq-mow-prop2}).
\qed

Careful: Note that $\lim_{\ap\rarrow a.}$
and $\sum_{a.}$ do
not commute. For example,
$\lim_{\ap\rarrow a.}\sum_{a.}\delta_{a.}^{\ap}=
1$ but
$\sum_{a.}\lim_{\ap\rarrow a.}\delta_{a.}^{\ap}=
\sum_{a.}1$.

\end{itemize}

\section{Do-Calculus}
As the notation for $P(b.|\myhat{a}.)$
suggests,
$P(b.|\myhat{a}.)$ and $P(b.|a.)$
are similar in some ways.
Recall that when
$P(b.|a.)$ is independent
of $a.$, we say that
$\rvb.$ is conditional independent of
$\rva.$.
Similarly, when
$P(b.|\myhat{a}.)$ is independent
of $\myhat{a}.$,
we might say that
$\rvb.$ is independent of
uprooting $\rva.$.
Furthermore,
conditioning on $\rva.$ and
uprooting $\rva.$
sometimes yield the same result.
The following theorem,
due to Pearl and Galles
(Ref.\cite{P95})
gives sufficient
graphical conditions
under which each of these 3 situations
will occur.

\begin{claim}(Do-Calculus Rules Theorem, Pearl and Galles):
Suppose $\rvx.$
is the set of all the nodes
of graph $G$
and $\rvx.$ equals the
union of the disjoint
subsets $\rva., \rvb.,\rvh.,\rvi.$
and $\rvo.$. (Note that in all the 3 rules
given below,
$\rvh.$ has a hat permanently
over it. That's why
I am using $\rvh$ for that variable,
as a mnemonic.)
\begin{itemize}
\item {\bf Rule 1 ($a.\leftrightarrow 1$)}:
\newline

\beq
(\rvb.\bot\rva.|\rvh.,\rvi.)
_{G_1}
\mbox{ where }
G_1 = G_{\myhat{\rvh}.}
\;
\eeq
iff, for all $b.,a.,h.,i.$,

\beq
P(b.:a.|\myhat{h}.,i.)=1
\;,
\eeq
or, equivalently,

\beq
P(b.|a.,\myhat{h}.,i.)=
P(b.|\myhat{h}.,i.)
\;.
\eeq

\item {\bf Rule 2 ($a.\leftrightarrow \;\;
\myhat{a}.$)}:

\beq
(\rvb.\bot\rva.|\rvh.,\rvi.)_{G_2}
\mbox{ where }
G_2 = G_{\myhat{\rvh}.,
\myvee{\rva}.}
\;
\eeq
iff, for all $b.,a.,h.,i.$,

\beq
P(b.:\myhat{a}.|
\myhat{h}.,i.)
=P(b.:a.|\myhat{h}.,i.)
\;,
\eeq
or, equivalently,

\beq
P(b.|\myhat{a}.,
\myhat{h}.,i.)=
P(b.|a.,\myhat{h}.,i.)
\;.
\eeq

\item {\bf Rule 3 ($\myhat{a}.
\leftrightarrow 1$)}:\newline
If

\beq
(\rvb.\bot\rva.|\rvh.,\rvi.)
_{G_3}
\;
\mbox{ where }
G_3 =
G_{\myhat{\rvh}.,
\left[\rva.-an(\rvi.,
G_{\myhat{\rvh}.})\right]^\wedge}
\;,
\eeq
then, for all $b.,a.,h.,i.$,

\beq
P(b.:\myhat{a}.|\myhat{h}.,i.)=1
\;,
\eeq
or, equivalently,

\beq
P(b.|\myhat{a}.,\myhat{h}.,i.)=
P(b.|\myhat{h}.,i.)
\;.
\label{eq-goal3}
\eeq
\end{itemize}

\end{claim}
\proof

The proofs presented below
for the do-calculus rules are
the same, except for some
minor modifications, as the proofs
first given by Pearl
(with
assistance from Galles for
the proof of rule 3) in the appendix of
Ref. \cite{P95}.
\begin{itemize}
\item{\bf Rule 1}:
By the D-sep Theorem,
$(\rvb.\bot\rva.|\rvh.,\rvi.)
_{G_1}$ iff

\beq
\delta_{\myhat{\rvh}.}
P(b.|a.,h.,i.)
=
\delta_{\myhat{\rvh}.}
P(b.|h.,i.)
\;.
\label{eq-sides1}
\eeq
If $LHS$ and $RHS$
denote the left and right hand sides
of Eq.(\ref{eq-sides1}), then

\beq
LHS=
P(b.|a.,\myhat{h}.,i.)
\;,
\eeq
and

\beq
RHS=
P(b.|\myhat{h}.,i.)
\;.
\eeq
\item{\bf Rule 2}: By the D-sep Theorem,
$(\rvb.\bot\rva.|\rvh.,\rvi.)
_{G_2}$ iff

\beq
\lim_{\ap\rarrow a.}
\delta_{\myvee{\rva}.(\ap)}
\delta_{\myhat{\rvh}.}
P(b.|a.,h.,i.)
=
\lim_{\ap\rarrow a.}
\delta_{\myvee{\rva}.(\ap)}
\delta_{\myhat{\rvh}.}
P(b.|h.,i.)
\;.
\label{eq-sides2}
\eeq
If $LHS$ and $RHS$
denote the left and right hand sides
of Eq.(\ref{eq-sides2}), then

\beq
LHS=
P(b.|a.,\myhat{h}.,i.)
\;,
\eeq
and

\beq
RHS=
P(b.|\myhat{a}.,\myhat{h}.,i.)
\;.
\eeq

\begin{figure}[h]
    \begin{center}
    \includegraphics[height=3.0in]{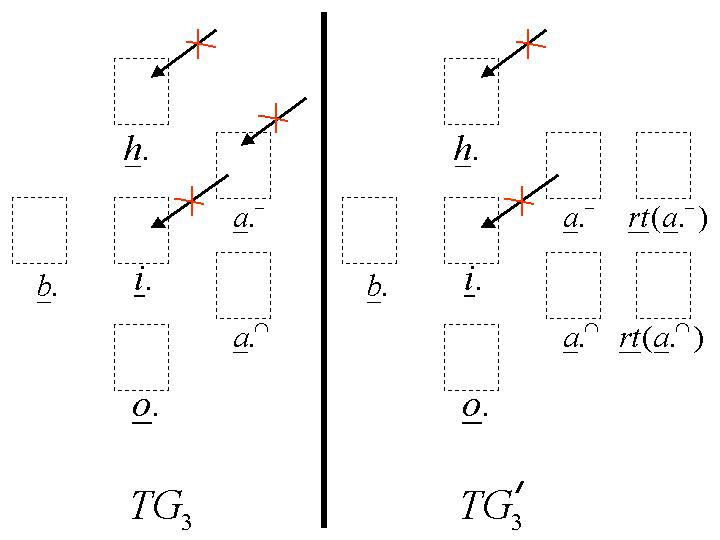}
    \caption{
    Graph templates
    $TG_3$ and $TG_3'$
    used in the proof of
    Rule 3 of do-calculus
    rules theorem.
    }
    \label{fig-rule-3}
    \end{center}
\end{figure}
\item{\bf Rule 3}:
Let
$\rva.^{-}$ and $\rva.^{\cap}$ be
abbreviations for the following sets of nodes:

\beq
\rva.^{-}=
\rva.-an(\rvi.,G_{\myhat{\rvh}.})
\;,
\eeq
and

\beq
\rva.^{\cap}=
\rva.\cap an(\rvi.,G_{\myhat{\rvh}.})
\;.
\eeq
Let ${\cal S}$ and ${\cal S}'$
denote the following statements

\beq
{\cal S}= (\rvb.\bot\rva.|\rvh.,\rvi.)
_{G_3}
\mbox{ where }
G_3 =
G_{\myhat{\rvh}.,
\left[\rva.^-\right]^\wedge}
\;,
\eeq
and

\beq
{\cal S}'=
(\rvb.\bot\rva., \rv{rt}(\rva.)|\rvh.,\rvi.)
_{G'_3}
\mbox{ where }
G_3'=
[G\larrow \rv{rt}(\rva.)]
_{\myhat{\rvh}.}
\;.
\eeq
By the D-sep Theorem,
${\cal S}'$ implies, for all
$b.,a.,rt(\rva.), h.,i.$,

\beq
P(b.|a.,rt(\rva.),\myhat{h}.,i.)=
P(b.|\myhat{h}.,i.)
\;.
\label{eq-pre-goal}
\eeq
But

\beq
P(b.|a.,rt(\rva.)=1,\myhat{h}.,i.) =
 P(b.|\myhat{a}.,\myhat{h}.,i.)
\;.
\eeq
So ${\cal S}'$
implies
Eq.(\ref{eq-goal3}).
Hence we'll be done with the proof if
we can prove that
${\cal S}$ implies ${\cal S}'$.
Let's prove this by proving
the contrapositive
$not({\cal S}')$ implies $not({\cal S})$.
If $not({\cal S}')$, then there
exists a path $\gamma$
which is unblocked at fixed
$\rvh. \rvi.$, and
which satisfies

\beq
\gamma \in Path_{TG_3'}(\rv{B}<\rv{A_1}<
\rv{A_2}<\ldots<\rv{A_n},\rv{rt}(\rv{A_n}))
\;,
\eeq
where $\rv{B}\in \rvb.$,
$\rv{A}.\subset \rva.$.
Here
$\rv{A_1}$
is the unique node in $\gamma$
that belongs to $\rva.$
and is closest to
$\rv{B}$.
But then there
is a shorter path
$\gamma_o$
in $TG_3'$
that is also
unblocked at fixed
$\rvh.,\rvi.$,

\beq
\gamma_o\in
Path_{T}(\rv{B}<\rv{A_1},\rv{rt}(\rv{A_1}))
\;,
\label{eq-gamma-o}
\eeq
where $T=TG_3'$.
If
we can show
that
$\gamma_o$
is unblocked at fixed $\rvh.,\rvi.$
and also satisfies
Eq.(\ref{eq-gamma-o})
with
$T=TG_3$
instead of the bigger set
$T=TG'_3$,
then we'll be done.
As shown
in Fig.\ref{fig-rule-3},
template $TG_3$
has the same
bans as template $TG'_3$
plus an additional ban
on arrows entering $\rva.^-$.
So we need to show that
$\gamma_o$ has
no arrows entering
$\rva.^-$.
Such an arrow would have
to enter node $\rv{A}_1$.
If
$\rv{A}_1\in \rva.^\cap$,
then there is no
arrow of $\gamma_o$ entering $\rva.^-$
and we are done.
If
$\rv{A}_1\in \rva.^-$,
then there are two possibilities,
either
$\rv{A}_1\in col(\gamma_o)$
or not.

If $\rv{A}_1\notin col(\gamma_o)$, since there
is an arrow pointing from $\rv{rt}(\rv{A}_1)$
to $\rv{A}_1$,
there must be an arrow pointing from
$\rv{A}_1$ to a node outside of $\rva.^-$.
Thus, there are no arrows in $\gamma_o$
entering $\rva.^-$ and we are done.

If $\rv{A}_1\in col(\gamma_o)$,
then, since $\gamma_o$
is unblocked at fixed $\rvh.,\rvi.$,
we must have $\overline{de}(\rv{A}_1)
\cap(\rvh.\cup\rvi.)\neq \emptyset$.
But this is impossible since
as shown by Fig.\ref{fig-rule-3},
in $TG_3$
no arrow can enter $\rvh.$
and no arrow from $\rva.^-$
can enter $\rvi.$.
\end{itemize}
.\qed

\end{document}